\documentclass{article}

\usepackage{arxiv}
\usepackage{hyperref}
\usepackage[utf8]{inputenc} 
\usepackage[T1]{fontenc}    
\usepackage{hyperref}       
\usepackage{url}            
\usepackage{booktabs}       
\usepackage{amsfonts}       
\usepackage{nicefrac}       
\usepackage{microtype}      
\usepackage{lipsum}		
\usepackage{graphicx}
\usepackage{natbib}
\usepackage{doi}
\usepackage{multirow}

\usepackage{subcaption}
\usepackage{xcolor}
\definecolor{RoyalBlue}{RGB}{65, 105, 225}

\title{Benchmark Assessment for DeepSpeed Optimization Library}

\author{ 
\href{https://orcid.org/0000-0002-6700-6664}{\includegraphics[scale=0.06]{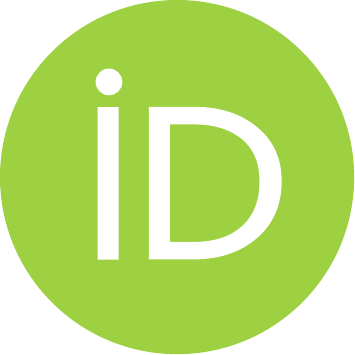}\hspace{1mm}Gongbo Liang} and 
\href{https://orcid.org/0000-0001-7832-5081}{\includegraphics[scale=0.06]{orcid.pdf}\hspace{1mm}Izzat Alsmadi}\\
	Department of Computing and Cyber Security\\
	Texas A\&M, San Antonio\\
	San Antonio, TX 78224 \\
	\texttt{tonybgky@gmail.com,  ialsmadi@tamusa.edu} \\
}

\renewcommand{\shorttitle}{\textit{Assessment for DeepSpeed Optimization Library} }

\hypersetup{
pdftitle={A template for the arxiv style},
pdfsubject={q-bio.NC, q-bio.QM},
pdfauthor={David S.~Hippocampus, Elias D.~Striatum},
pdfkeywords={First keyword, Second keyword, More},
}

\begin{document}
\maketitle

\begin{abstract}


Deep Learning (DL) models are widely used in machine learning due to their performance and ability to deal with large datasets while producing high accuracy and performance metrics. The size of such datasets and the complexity of DL models cause such models to be complex, consuming large amount of resources and time to train. Many recent libraries and applications are introduced to deal with DL complexity and efficiency issues. In this paper, we evaluated one example, Microsoft DeepSpeed library through classification tasks. DeepSpeed public sources reported classification performance metrics on the LeNet architecture. We extended this through evaluating the library on several modern neural network architectures, including convolutional neural networks (CNNs) and Vision Transformer (ViT). Results indicated that DeepSpeed, while can make improvements in some of those cases, it has no or negative impact on others.

\end{abstract}

\keywords{Machine Learning \and Neural Networks \and Deep Learning Models \and Optimization Models }

\section{Introduction}

The popularity of Neural Network and Deep Learning models is unprecedented. There are the prime choice for many machine learning models and applications. In comparison with traditional machine learning algorithms, Deep Learning algorithms perform better when datasets are large, \cite{mahapatra2018deep}.
However, the number of proposed neural network models and algorithms is large. In addition, these models can be trained with a wide range of hyperparameters, including the choice of learning rate, optimizer, length of training (i.e., number of epochs), etc. Finding the optimal model and hyperparameter for a particular dataset may not be straightforward without extensive evaluations. Besides, the complexity of models has been kept increasing due to the positive impact of complexity, such as including more layers may lead to a better accuracy. However, due to the large size and requirement for tremendous computational resources, such complex models may not be practical for many applications and limit their usage in production environments. 

Large-scale distributed training improves the performance of training large and complex models. In those approaches, data parallelism is adopted and SGD is usually selected as the optimization method because of its high computation
efficiency and well support by the DL tool-kits, such as TensorFlow, PyTorch and DeepSpeed~\cite{he2021large}. In SGD, each worker processes a random mini-batch of training data. Quantized or
sparcified SGD allows each worker to use fewer bits to pass gradients by sacrificing the convergence to a mild extent.

There are four stages during each in DL data parallel training, \cite{wang2021lightseq2}:
 \begin{itemize}
     \item The model receives input data and then performs forward propagation.
\item The model performs backward propagation using the
loss calculated after forward propagation, generating
the parameters gradients.
\item The averaged gradients are computed and broadcast to each
device. 
\item All parameters in each device are updated using the
averaged gradients. 
 \end{itemize}

\subsection{Distributed Deep Learning Frameworks}
Besides, DeepSpeed, there are several other examples of distributed deep learning frameworks.These frameworks provide structured methods to define DL models using a collection of pre-built and pre-optimized components. 
\begin{itemize}
\item TensorFlow, \cite{abadi2016tensorflow}.
\item Caffe and MPI Caffe, \cite{vision2019caffe}, \cite{jia2014caffe}.
\item Keras, \cite{chollet2018keras}.
\item PyTorch DDP
\item Microsoft Computational Network Toolkit, CNTK
\item BigDL, \cite{dai2019bigdl}.
\item SINGA, \cite{ooi2015singa}.
\item MXNET-MPI, \cite{chen2015mxnet}.
\item Horovod, \cite{sergeev2018horovod}.
\item DeepSpeed, \cite{Rasley2020DeepSpeed}. 

\end{itemize}

\section{Related Work}
\subsection{Microsoft DeepSpeed: Evaluate models with a large number of parameters}
DeepSpeed is a recent DL library made available to public by Microsoft in 2020. "DeepSpeed brings state-of-the-art training techniques, such as Zero Redundancy Optimizer (ZeRO): a novel memory optimization technology for large-scale distributed DL, optimized kernels, distributed training, mixed precision, and checkpointing, through lightweight APIs compatible with PyTorch", \cite{Rasley2020DeepSpeed}. DeepSpeed is designed to enable a large set of input parameters. It is optimized for low latency and high throughput training, \cite{sedonascalable2020}. DeepSpeed
implementation of 3D parallelism can scale to over a trillion parameters on 800 NVIDIA V100 GPUs by fully leveraging the aggregate
GPU memory of a cluster, \cite{branwen2019september}.
The library is compatible with PyTorch in its current version. Examples of implementations of this library can be evaluated through \url{https://github.com/microsoft/DeepSpeedExamples}. In one example using BERT training dataset, the library showed a 34\% efficiency improvement over the best published results. 
DeepSpeed main functions can be summarized into 3 major components:

\begin{itemize}
    \item Zero:  Zero, introduced by \cite{rajbhandari2020zero} optimizes memory, which can improve training speed while increasing the model size that can be efficiently trained. This parallelized optimizer greatly reduces the resources needed for parallelism while increasing the number of parameters that can be trained, \cite{Rasley2020DeepSpeed}. Data parallelism is not new in general as its already employed by High Performance Computing (HPC) systems that can employ parallelism not only in data, but also in the software, hardware and also the network.
    \item  Sparse Attention: Attention-based DL models, such as
Transformers, are effective in capturing relationships between input tokens. Sparse attention is an improvement of the attention mechanism to extract patterns from sequences longer than possible previously. In Transformers, it gets impractical to compute a single attention matrix, for very large inputs. Accordingly, and as an alternative to full attention, in sparse attention, each output position only computes weightings from a subset of input positions. Research papers showed that t sparse attention is sufficient to get state-of-the-art results in modeling long sequences over language modeling, image
generation and music generation, \cite{sukhbaatar2019adaptive}.
    \item  1-bit Adam: 
    State-of-the-art error compensation techniques only work with basic optimizers such as SGD, which are linearly dependent on the gradients. They do not work with non-linear gradient-based optimizers such as Adam, \cite{tang20211}.  1-bit Adam is proposed to reduce the communication volume by up to 5×, offers better scalability, and provides the same sample-wise convergence speed as uncompressed Adam, \cite{tang20211}.
\end{itemize}  

One limitation with DeepSpeed is that it can only support Transformer encoder layer, thus can only be used to train BERT-like models, \cite{devlin2018bert}. DeepSpeed also requires the model to fit across the combined memory of all the GPU devices, \cite{pudipeddi2020training}.

\subsection{Models optimization and over-parameterized systems}

DL models, often consist of huge amounts of parameters that far
exceed the instance numbers, \cite{bassily2018exponential}, \cite{ma2018power}, \cite{allen2019convergence}, \cite{oymak2019overparameterized}.
Over-parameterization in DL networks has been shown to have advantages, where researchers indicated that wider networks train faster and have better generalization performance \cite{neyshabur2014search}, \cite{arpit2017closer}.
Deep learning models are effective due to the effectiveness of local gradient-based optimization methods, such as Stochastic Gradient Descent (SGD), in training large neural networks, \cite{liu2020toward}.  Several publications (e.g. \cite{jacot2018neural}) connected effective optimization of over-parameterized networks to properties of their linearizations.

\section{Experiments and Discussion}
This project tests seven neural network architectures, two optimizers, and two learning rate scheduling methods using the CIFAR-10 dataset~\cite{krizhevsky2009learning}. In total, 42 training trails of 21 unique combinations of the architecture, optimizer, and learning rate scheduling method are tested. Each unique combination is trained twice, with and without DeepSpeed, respectively.

We compare the performance between corresponding training trails in terms of classification accuracy and the training efficiency (i.e., seconds per epoch training). We start this section by giving an introduction to the experimental setup (Section~\ref{sec:setup}). The detailed evaluation results are presented in Sections~\ref{sec:acc} and~\ref{sec:run_time}. Finally, this section ends with the discussion and limitation in Sections~\ref{sec:limit}. 

\subsection{Experimental Setup}
\label{sec:setup}
The Vision Transformer (ViT)~\cite{dosovitskiy2020vit} and six CNN arechitectures---namely LeNet~\cite{lecun1998gradient}, AlexNet~\cite{krizhevsky2012imagenet}, VGG11\_BN~\cite{simonyan2014very}, ResNet-18~\cite{he2016deep}, DenseNet-121~\cite{huang2017densely}, and SequeezeNet-v1.0~\cite{iandola2016squeezenet}--- are tested on image classification tasks using CIFAR-10. The LeNet is implemented by ourselves, the ViT model is loaded from the timm library~\cite{rw2019timm}, and the six CNNs are loaded from PyTorch. For all of the networks, we modify the output size of the last layer to be ten to match the number of classes in CIFAR-10. The input images are scaled to the proper size to match the input layer size of each model, ranging between $32\times32$ and $299\times299$.

Two optimizers, SGD and Adam~\cite{kingma2014adam}, and two learning rate scheduling methods (i.e., with or without learning rate scheduler) are also evaluated in this study. The WarmupLR learning rate scheduler is used for DeepSpeed models when a learning rate scheduler is used, and the CyclicLR~\cite{smith2017cyclical} is used for the normal models (i.e., without using DeepSpeed) when a learning rate scheduler is used. The WarmupLR is adopted from the DeepSpeed examples\footnote{https://github.com/microsoft/DeepSpeedExamples}. However, since PyTorch does not provide an implementation for WarmupLR, we decide to use CyclicLR instead. The base learning rate is set to $10^{-3}$ for all the training trials. If a learning rate scheduler is used, we set the minimum learning rate as $0$ and maximum learning rate as $10^{-3}$. 

All the experiments are conducted using a Linux workstation with two Nvidia 3090 GPU cards, with a 24 GB memory for each of them. The DeepSpeed training models are distributed to both of the GPUs, and the normal training models are trained using only one GPU card. Each model is trained for 25 epochs from scratch with a batch size of 16. Cross-entropy loss is used during the training. In addition, half-precision (FP16) is used for DeepSpeed training, and full precision (FP32) is used for normal training.

\subsection{Experimental Results and Analysis}

\subsubsection{Classification Accuracy}
\label{sec:acc}
With the architectures, optimizers, and learning rate scheduling methods mentioned above, 21 permutations are derived. We train each permutation twice, with or without DeepSpeed. Table~\ref{table:cifar10_acc} shows the best performance of 42 trained model. The words \textbf{SGD} and \textbf{Adam} indicate the optimizer used in the model training. The postfix \textbf{+S} shows whether a learning rate scheduler is used in training. The highest accuracy of each neural network architecture is highlighted in bold. The table reveals the best performance of different architectures is often coming from normal training.

\begin{table}[!tb]
	\centering
	\caption{Classification Performance of Different Models}
	\begin{tabular}{lcccccc}
	\hline \hline
	\noalign{\smallskip}
	\multirow{2}{*}{\textbf{Models}} & \multicolumn{3}{c}{\textbf{DeepSpeed Training}} & \multicolumn{3}{c}{\textbf{Normal Training}} \\ 
	& \textbf{  SGD  } & \textbf{SGD+S} & \textbf{Adam+S} & \textbf{  SGD  } & \textbf{SGD+S} & \textbf{Adam+S}\\ \hline
    \noalign{\smallskip}
    
    \textbf{LeNet} & 58.41 & 57.15 & 58.37 & \textbf{65.62} & 64.13 & 64.95 \\ \noalign{\smallskip}
    \textbf{AlexNet} & 60.39 & 61.64 & 62.56 & \textbf{82.60} & 80.54 & 70.61 \\ \noalign{\smallskip}
    \textbf{VGG11\_BN} & 66.26 & 66.11 & 10.28 & \textbf{81.00} & 79.62 & 78.04  \\ \noalign{\smallskip}
    \textbf{ResNet-18} & 56.84 & 57.55 & 79.72 & 73.46 & 70.43 & \textbf{80.69} \\ \noalign{\smallskip}
    \textbf{DenseNet-121} & 52.70 & 53.06 & 80.70 & 75.61 & 70.86 & \textbf{84.00} \\ \noalign{\smallskip}
    \textbf{SqueezeNet-v1.0} & 55.55 & 55.24 & 73.48 & 75.79 & 72.06 & \textbf{78.91} \\ \noalign{\smallskip}
    \textbf{ViT} & 58.87 & 59.40 & 13.15 & 61.22 & \textbf{63.45} & 20.20 \\ \noalign{\smallskip}
    \hline \hline
  \end{tabular}
  \label{table:cifar10_acc}
\end{table}

\begin{figure}[!tb]
     \centering
     \begin{subfigure}[b]{0.485\textwidth}
         \centering
         \includegraphics[width=\textwidth]{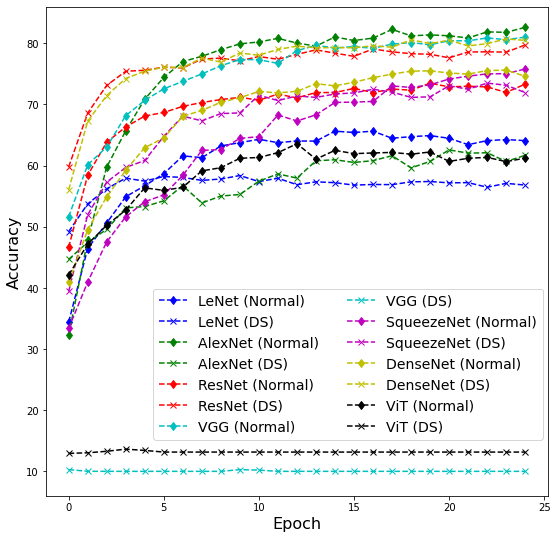}
         \caption{original setting}
         \label{fig:cifar10_original}
     \end{subfigure}~~~
     \begin{subfigure}[b]{0.485\textwidth}
         \centering
         \includegraphics[width=\textwidth]{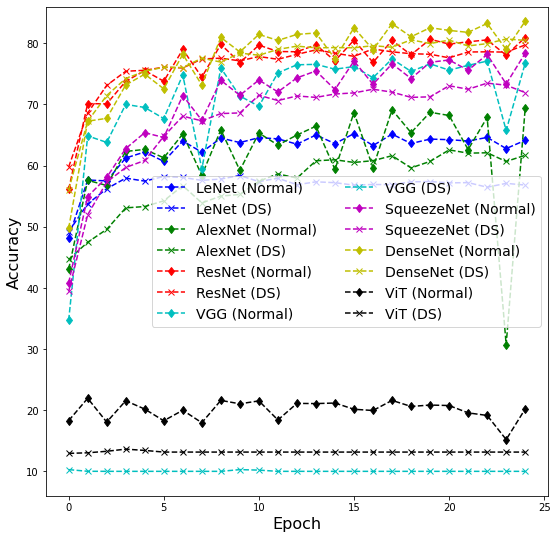}
         \caption{adam + cycliclr}
         \label{fig:cifar10_adam_scheduler}
     \end{subfigure}
     
     \begin{subfigure}[b]{0.485\textwidth}
         \centering
         \includegraphics[width=\textwidth]{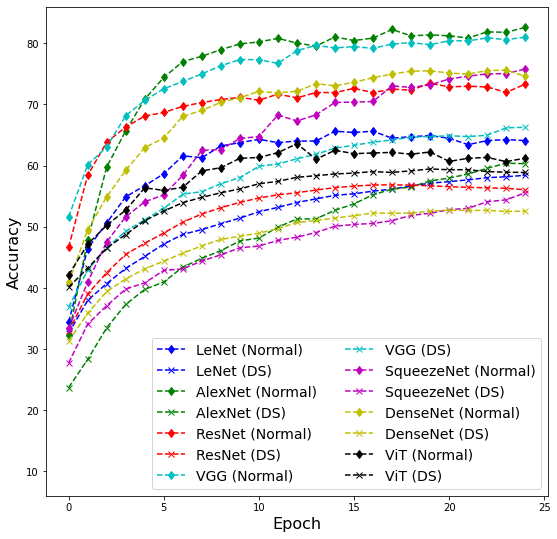}
         \caption{sgd + no scheduler}
         \label{fig:cifar10_sdg}
     \end{subfigure}~~~
     \begin{subfigure}[b]{0.485\textwidth}
         \centering
         \includegraphics[width=\textwidth]{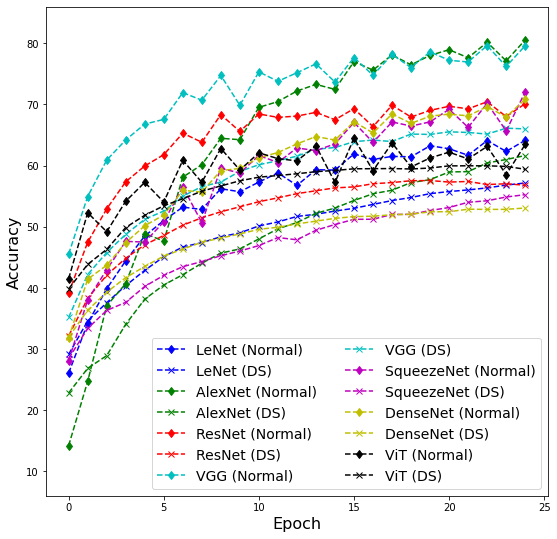}
         \caption{sgd + cycliclr}
         \label{fig:cifar10_sdg_scheduler}
     \end{subfigure}
     
    \caption{}
    \label{fig:cifar10_acc}
\end{figure}

To better analyze the result, we group the models into four different settings and present them in four plots: 
\begin{itemize}
    \item I. Adam optimizer with learning rate scheduler for DeepSpeed models, and uses SGD without scheduler for normal models (Figure~\ref{fig:cifar10_original});
    \item II. Adam optimizer with learning rate scheduler for all models (Figure~\ref{fig:cifar10_adam_scheduler}); 
    \item III. SGD optimizer without learning rate scheduler for all models (Figure~\ref{fig:cifar10_sdg}); 
    \item IV. SGD optimizer with learning rate scheduler  for all models (Figure~\ref{fig:cifar10_sdg_scheduler}).
\end{itemize}

Figure~\ref{fig:cifar10_original} shows the results of the models trained using Setting I, which is the original setting in the DeepSpeed CIFAR-10 example\footnote{https://github.com/microsoft/DeepSpeedExamples/tree/master/cifar}. We follow the example almost exactly, except we include six more neural network architectures. The figure shows that the DeepSpeed training models normally perform better when the training epoch is small, such as 1 or 2, except the VGG model. However, when the training epoch increases, the difference between normal and DeepSpeed training reduces. The normal training models might have a significant performance improvement than the corresponding counterpart over time, especially when the model compacity is small, such as LeNet and AlexNet. However, for models with larger compacity, such as ResNet and DenseNet, the performance of DeepSpeed training models are consistently better than the normal training models. It seems the experiment suggests DeepSpeed may be beneficial when the model capacity is larger. However, the comparison between the DeepSpeed and normal models could be biased because different optimizes were used in the example, especially given the fact that Adam optimizer is known for achieving a better performance in a shorter time frame.

To create a more fair comparison, we try to train both methods using the same hyperparameters or similar hyperparameters. Figure~\ref{fig:cifar10_sdg} shows the performance of models trained with SGD optimizer only (i.e., Setting III). No learning rate scheduler is used. The figure shows that all normal training models perform better than the corresponding DeepSpeed models. In addition, the performance difference between the two training methods (i.e., with or without DeepSpeed) could be as large as more than 20\% for a particular architecture. For instance, DenseNet with DeepSpeed training is 52.70\% accuracy. However, the performance of DenseNet with normal training is 75.61\%.

Figure~\ref{fig:cifar10_sdg_scheduler} shows the performance of models trained using SGD optimizer and learning rate scheduler (Setting IV). The result is very close to Figure~\ref{fig:cifar10_sdg}, which uses SGD optimizer only. The WarmupLR scheduler does improve the performance of DeepSpeed models. However, the improvement is marginal, with an average of less than 1\%. In general, DeepSpeed models still have significantly worse performance than normal training models.

Figure~\ref{fig:cifar10_adam_scheduler} shows the models trained using Adam optimizer with learning rate scheduler (Setting II). The figure reveals that the DeepSpeed models' performance improved dramatically when using Adam optimizer. For instance, the performance of DeepSpeed trained ResNet is improved from about 57\% accuracy to close to 80\%, compared with SGD optimizer. However, this performance improvement is not limited to DeepSpeed models. The normal training ResNet model is also improved from lower 70\% accuracy to 80\%. Similarly, with SGD settings, the normal training models have better performance than the DeepSpeed training models.

\subsubsection{Training Efficiency}
\label{sec:run_time}

Table~\ref{table:run_time} shows the training efficiency of each model with or without using DeepSpeed. We report the training efficiency in terms of seconds per epoch to train. For each pair of training, the one trained with less time is highlighted in bold. The table shows that normal training is faster for almost all of the convolutional neural network models, except DenseNet-121 with Adam optimizer. In addition, for almost all the convolutional neural network models, DeepSpeed training may take a significantly longer time, especially for AlexNet, the training time event got almost tripled. However, when evaluating the training time of ViT models, DeepSpeed shows a clear advantage, which usually takes only 2/3 of the training time compared with normal training.

\begin{table}[!tb]
	\centering
	\caption{Training Time of Different Models in Seconds per Epoch}
	\begin{tabular}{l c c c c}
	\hline \hline
	\noalign{\smallskip}
	
    \textbf{Models} & \textbf{Optimizer} & \textbf{Use Scheduler} &\textbf{DeepSpeed} & \textbf{Normal} \\ \noalign{\smallskip} \hline
    
    \multirow{3}{*}{\textbf{LeNet}} & \multirow{2}{*}{SGD}  & No & 16.63 & \textbf{11.97} \\
     &  & Yes & 16.72 & \textbf{12.05} \\
     & Adam & Yes & 14.95 & \textbf{13.32} \\ \hline
    
    \multirow{3}{*}{\textbf{AlexNet}} & \multirow{2}{*}{SGD}  & No & 101.41 & \textbf{35.72} \\
     &  & Yes & 101.46 & \textbf{35.56} \\
     & Adam & Yes & 97.95 & \textbf{48.61} \\ \hline
    
    \multirow{3}{*}{\textbf{VGG11\_BN}} & \multirow{2}{*}{SGD}  & No & 231.62 & \textbf{156.71} \\
     &  & Yes & 231.57 & \textbf{156.90} \\
     & Adam & Yes & 219.27 & \textbf{185.93} \\ \hline
     
    \multirow{3}{*}{\textbf{ResNet-18}} & \multirow{2}{*}{SGD}  & No & 97.18 & \textbf{66.08} \\
     &  & Yes & 99.34 & \textbf{66.04} \\
     & Adam & Yes & 86.55 & \textbf{73.46} \\ \hline 
    
    \multirow{3}{*}{\textbf{DenseNet-121}} & \multirow{2}{*}{SGD}  & No & 346.94 & \textbf{273.53} \\
     &  & Yes & 348.22 & \textbf{275.50} \\
     & Adam & Yes & \textbf{288.87} & 318.77 \\ \hline 
     
    \multirow{3}{*}{\textbf{SqueezeNet-v1.0}} & \multirow{2}{*}{SGD}  & No & 83.86 & \textbf{61.40} \\
     &  & Yes & 85.43 & \textbf{61.47} \\
     & Adam & Yes & 73.41 & \textbf{64.80} \\ \hline
     
    \multirow{3}{*}{\textbf{ViT}} & \multirow{2}{*}{SGD}  & No & \textbf{259.05} & 354.91 \\
     &  & Yes & \textbf{259.56} & 384.52 \\
     & Adam & Yes & \textbf{225.54} & 376.50 \\
    \hline \hline
  \end{tabular}
  \label{table:run_time}
\end{table}

\subsection{Discussions and Limitations}
\label{sec:limit}
The primary goal of this study is to evaluate the DeepSpeed library performance on classification models by extending the official example\footnote{https://github.com/microsoft/DeepSpeedExamples/tree/master/cifar}. Initially, we want to want to use the example as-is to reduce the possibility of introducing bias to the experiences. However, we noticed that the official example uses different optimizers and learning rate scheduling methods to train the DeepSpeed and normal models. More specifically, the DeepSpeed model uses the Adam optimizer with a learning rate scheduler, and the normal model uses the SGD optimizer without a learning rate scheduler. Thus, we decided to expand our experiments to cover both Adam and SGD optimizers, as well as with and without a learning rate scheduler. However, our experiments may still be biased on some hyperparameters. 
For instance, all the models are initialized using random seedings, which may introduce uncertainty to each training trial. The DeepSpeed models are trained by distributing to two GPU cards for training; however, the normal models are trained using only one GPU card. Half-precision (FP16) is used for DeepSpeed training, and full precision (FP32) is used for normal training.

According to our experimental result, the DeepSpeed library does not be very helpful for CNNs, but it works well for ViT. This makes us wonder if the library is designed only for neural networks beyond the normal training scale. In addition, the official Github site\footnote{https://github.com/microsoft/DeepSpeed} of DeepSpeed claims to make distributed training easy, efficient, and effective. It may train 10x larger models and provide 10x faster training. Two particular observations from our experimental result may support this hypothesis.

For instance, Table~\ref{table:run_time} shows DeepSpeed reduces the training time of ViT models by up to 40\%. However, it does not help for most CNN models. This result may show models with larger capacity may be benefited more from using DeepSpeed. In addition, if we compare the ViT classification performance against all other neural network architectures that trained using the same method, a bigger performance improvement is observed when using DeepSpeed. For instance, Table~\ref{table:cifar10_acc} shows that without using DeepSpeed, ViT has the worst performance regardless of the optimizer and scheduler. However, with using DeepSpeed, ViT performed the $3^{rd}$ best amount of all the tested architectures when using SGD as the optimizer. This observation may also support our hypothesis that the DeepSpeed library is designed only for neural networks beyond the normal training scale. Thus, we would focus more on evaluating DeepSpeed on models with larger capacity in the future.

\section{Summary and Conclusion}
In this paper, we evaluated the performance of the DeepSpeed library on classification tasks across seven neural network architectures, two optimizers, and two learning rate scheduling methods. \\
Our results demonstrate that DeepSpeed may improve model performance while reducing training time in cases where the model capacity is large. However, it has no or negative impact on those models with smaller capacity. It is important to mention, however, that our experiment may be limited by using a toy dataset (i.e., CIFAR-10) and some other factors. We are planning to expand our study to cross multiple datasets and more focus on models with larger capacity in terms of images, etc.

\bibliographystyle{unsrtnat}
\bibliography{Benchmark_Assessment_DS_Lib}  






\end{document}